\documentclass[onecolumn,journal,onecolumn]{IEEEtran}
\long\def\twocolumn[#1]{#1}

\makeatletter\if@twocolumn\PassOptionsToPackage{switch}{lineno}\else\fi\makeatother

\usepackage{graphicx}
\usepackage[T1]{fontenc}
\ifCLASSINFOpdf
\else
\fi
\usepackage{amsmath}
\usepackage[cmintegrals]{newtxmath}
\usepackage{url,multirow,morefloats,floatflt,cancel,tfrupee}
\makeatletter

\AtBeginDocument{\@ifpackageloaded{textcomp}{}{\usepackage{textcomp}}}
\makeatother
\usepackage{colortbl}
\usepackage{xcolor}
\usepackage{pifont}
\usepackage[nointegrals]{wasysym}
\makeatletter

\def\mcWidth#1{\csname TY@F#1\endcsname+\tabcolsep}

\def\cAlignHack{\rightskip\@flushglue\leftskip\@flushglue\parindent\z@\parfillskip\z@skip}
\def\rAlignHack{\rightskip\z@skip\leftskip\@flushglue \parindent\z@\parfillskip\z@skip}

\usepackage{ifxetex}
\ifxetex\else\if@twocolumn\usepackage{dblfloatfix}\fi\fi

\AtBeginDocument{
\expandafter\ifx\csname eqalign\endcsname\relax
\def\eqalign#1{\null\vcenter{\def\\{\cr}\openup\jot\m@th
  \ialign{\strut$\displaystyle{##}$\hfil&$\displaystyle{{}##}$\hfil
      \crcr#1\crcr}}\,}
\fi
}

\AtBeginDocument{%
  \@ifpackageloaded{endfloat}%
   {\renewcommand\efloat@iwrite[1]{\immediate\expandafter\protected@write\csname efloat@post#1\endcsname{}}}{}%
}%

\def\BreakURLText#1{\@tfor\brk@tempa:=#1\do{\brk@tempa\hskip0pt}}
\let\lt=<
\let\gt=>
\def\processVert{\ifmmode|\else\textbar\fi}

\@ifundefined{subparagraph}{
\def\subparagraph{\@startsection{paragraph}{5}{2\parindent}{0ex plus 0.1ex minus 0.1ex}%
{0ex}{\normalfont\small\itshape}}%
}{}

\newcommand\role[1]{\unskip}
\newcommand\aucollab[1]{\unskip}
  
\@ifundefined{tsGraphicsScaleX}{\gdef\tsGraphicsScaleX{1}}{}
\@ifundefined{tsGraphicsScaleY}{\gdef\tsGraphicsScaleY{.9}}{}
\def\checkGraphicsWidth{\ifdim\Gin@nat@width>\linewidth
	\tsGraphicsScaleX\linewidth\else\Gin@nat@width\fi}

\def\checkGraphicsHeight{\ifdim\Gin@nat@height>.9\textheight
	\tsGraphicsScaleY\textheight\else\Gin@nat@height\fi}

\def\fixFloatSize#1{}
\let\ts@includegraphics\includegraphics

\def\inlinegraphic[#1]#2{{\edef\@tempa{#1}\edef\baseline@shift{\ifx\@tempa\@empty0\else#1\fi}\edef\tempZ{\the\numexpr(\numexpr(\baseline@shift*\f@size/100))}\protect\raisebox{\tempZ pt}{\ts@includegraphics{#2}}}}

\AtBeginDocument{\def\includegraphics{\@ifnextchar[{\ts@includegraphics}{\ts@includegraphics[width=\checkGraphicsWidth,height=\checkGraphicsHeight,keepaspectratio]}}}

\DeclareMathAlphabet{\mathpzc}{OT1}{pzc}{m}{it}

\def\URL#1#2{\@ifundefined{href}{#2}{\href{#1}{#2}}}

\def\UrlOrds{\do\*\do\-\do\~\do\'\do\"\do\-}%
\g@addto@macro{\UrlBreaks}{\UrlOrds}

\@ifundefined{quoteAttrib}
	{}
	{}

\@ifundefined{titlequoteAttrib}
	{}{}

\newenvironment{title-quote}
	{\list{}{\fontsize{10pt}{12pt}\selectfont\leftmargin.5in\itshape\rightmargin\leftmargin}%
  \item\relax}
  {\endlist}

\makeatother

\def\style#1#2{#2}

\usepackage{tabulary}
\makeatletter
\AtBeginDocument{\@ifpackageloaded{longtable}{%
\def\LT@makecaption#1#2#3{%
  \LT@mcol\LT@cols c{\hbox to\z@{\hss\parbox[t]\LTcapwidth{%
    \sbox\@tempboxa{#1{#2: } #3}%
    \ifdim\wd\@tempboxa>\hsize
      #1{#2: }\textsc{#3}%
    \else
      \hbox to\hsize{\hfil\box\@tempboxa\hfil}%
    \fi
    \endgraf\vskip\baselineskip}%
  \hss}}}
}{}}
\makeatother

   \makeatletter
  \def\fig@textbf{\textbf}
   \AtBeginDocument{\renewcommand\floatc@plain[2]{\setbox\@tempboxa\hbox{{\footnotesize#1.}\footnotesize\hskip.5em#2}%
    \ifdim\wd\@tempboxa>\hsize {\fig@textbf{\footnotesize#1.}}\footnotesize\hskip.5em#2\par
        \else\hbox to\hsize{\hfil\box\@tempboxa\hfil}\fi}}
    \makeatother
  
\usepackage{float}

\begin{document}

%

        \title{Activation Functions: Comparison of Trends in Practice and Research for Deep Learning }
      
\author{Chigozie~Enyinna~Nwankpa,
        Winifred~Ijomah,
        Anthony~Gachagan, and 
        Stephen~Marshall\thanks{Chigozie Enyinna~Nwankpa, Winifred~Ijomah are with Design Manufacturing and Engineering Management, Faculty of Engineering, University of Strathclyde, Glasgow, UK, e-mail: chigozie.nwankpa@strath.ac.uk, e-mail: w.l.ijomah@strath.ac.uk}\thanks{Anthony~Gachagan, Stephen~Marshall are with Electronic and Electrical Engineering, Faculty of Engineering, University of Strathclyde, Glasgow, UK, e-mail: a.gachagan@strath.ac.uk, e-mail: stephen.marshall@strath.ac.uk}}

\maketitle 

\begin{abstract}
Deep neural networks have been successfully used in diverse emerging domains to solve real world complex problems with may more deep learning(DL) architectures, being developed to date. To achieve these state-of-the-art performances, the DL architectures use activation functions (AFs), to perform diverse computations between the hidden layers and the output layers of any given DL architecture.

This paper presents a survey on the existing AFs used in deep learning applications and highlights the recent trends in the use of the activation functions for deep learning applications.  The novelty of this paper is that it compiles majority of the AFs used in DL and outlines the current trends in the applications and usage of these functions in practical deep learning deployments against the state-of-the-art research results. This compilation will aid in making effective decisions in the choice of the most suitable and appropriate activation function for any given application, ready for deployment.

This paper is timely because most research papers on AF highlights similar works and results while this paper will be the first, to compile the trends in AF applications in practice against the research results from literature, found in deep learning research to date.

\end{abstract}
    

%
\IEEEpeerreviewmaketitle

\section{Introduction}
Deep learning algorithms are multi-level representation learning techniques that allows simple non-linear modules to transform representations from the raw input into the higher levels of abstract representations, with many of these transformations producing learned complex functions. The deep learning research was inspired by the limitations of the conventional learning algorithms especially being limited to processing data in raw form, \unskip~\cite{296343:6576746} and the human learning techniques by changing the weights of the simulated neural connections on the basis of experiences, obtained from past data \unskip~\cite{296343:6577200}.

The use of representation learning, which is the technique that allow machines to discover relationships from raw data, needed to perform certain tasks likes classification and detection. Deep learning, a subfield of machine learning, is more recently being referred to as representation learning in some literature\unskip~\cite{296343:6576722}. The direct relationships between deep learning and her associated fields can be shown using the relationship Venn diagram in Figure~\ref{figure-dcef9a16296475181e39945203c8f5e9}.

\bgroup
\fixFloatSize{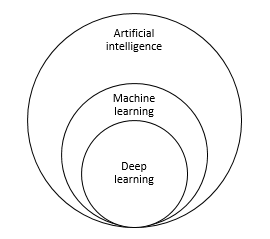}
\begin{figure}[!htbp]
\centering \makeatletter\IfFileExists{images/05fac13b-c4fa-4557-b6dd-651402b0fa71-udeeplearning.PNG}{\includegraphics{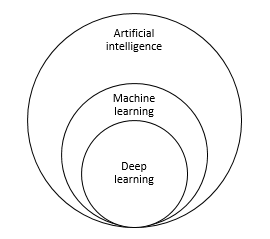}}{}
\makeatother 
\caption{{Venn diagram of the components of artificial intelligence}}
\label{figure-dcef9a16296475181e39945203c8f5e9}
\end{figure}
\egroup
Over the past six decades, machine learning field, a branch of artificial intelligence started rapid expansion and research in the field gained momentum to diversify into different aspects of human existence. Machine learning is a field of study that uses the statistics and computer science principles, to create statistical models, used to perform major tasks like predictions and inference. These models are sets of mathematical relationships between the inputs and outputs of a given system. The learning process is the process of estimating the models parameters such that the model can perform the specified task\unskip~\cite{296343:6576710}. For machines, the learning process tries to give machines the ability to learn without being programmed explicitly.

The typical artificial neural networks (ANN) are biologically inspired computer programmes, designed by the inspiration of the workings of the human brain. These ANNs are called networks because they are composed of different functions\unskip~\cite{296343:7284218}, which gathers knowledge by detecting the relationships and patterns in data using past experiences known as training examples in most literature. The learned patterns in data are modified by an appropriate activation function and presented as the output of the neuron as shown in Figure~\ref{figure-8d58b6b37eba86ad2e86a2e491a65c69}

\bgroup
\fixFloatSize{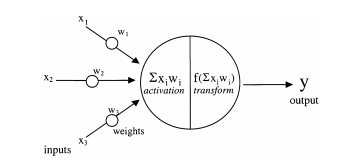}
\begin{figure}[!htbp]
\centering \makeatletter\IfFileExists{images/4d446cfe-c6b4-4827-ba55-7706b4d83358-uneurons.PNG}{\includegraphics{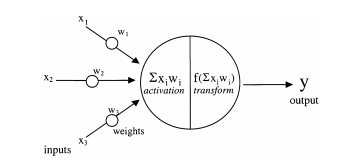}}{}
\makeatother 
\caption{{Typical biological inspired neuron\unskip~\protect\cite{296343:6576708}}}
\label{figure-8d58b6b37eba86ad2e86a2e491a65c69}
\end{figure}
\egroup
The early deep learning algorithms used for recognition tasks had few layers in their entire architecture, with LeNet5, having just five layers\unskip~\cite{296343:6576747}. These network layers have witnessed depth increases since then, with AlexNet having twelve layers\unskip~\cite{296343:6576743}, VGGNet having sixteen and nineteen layers in its two variants\unskip~\cite{296343:6576761}, twenty-two layers in GoogleNet\unskip~\cite{296343:6577207}, one hundred and fifty-two layers in the largest ResNet architecture\unskip~\cite{296343:6576734} and over one thousand two hundred layers in Stochastic Depth networks, already trained successfully\unskip~\cite{296343:6577209}, with the layers still increasing to date. With the networks getting deeper, the need to understand the makeup of the hidden layers and the successive actions taking place within the layers becomes inevitable.

However, the major issue of using deep neural network architectures is the difficulty of developing the algorithms to effectively learn the patterns in data and studies on these issues associated with the training of neural networks, have been a key research area to date. 

To improve the performance of these learning algorithms, Turian et al., 2009 highlighted that three key areas are usually of interest namely: to improve the model, to engineer better features, and to improve inference\unskip~\cite{296343:6576765}. The use of better feature models has worked in many deep learning applications to obtain good model feature for the applications alongside the improvement of models but the key interest area lies in inference improvement of models for deep learning applications.

Several techniques for model improvement for deep learning algorithms exist in literature which include the use of batch-normalization and regularization,\unskip~\cite{296343:6636741,296343:6636743,296343:6636739,296343:6636740}, dropout\unskip~\cite{296343:6636741}, proper initialization\unskip~\cite{296343:7519235}, good choice of activation function to mention a few\unskip~\cite{296343:6636743,296343:7519235,296343:6576757,296343:6576727}. These different techniques offer one form of improvement in results or training improvement but our interest lies in the activation functions used in deep learning algorithms, their applications and the benefits of each function introduced in literature.

A common problem for most learning based systems is, how the gradient flows within the network, owing to the fact that some gradients are sharp in specific directions and slow or even zero in some other directions thereby creating a problem for an optimal selection techniques of the learning parameters. 

The gradients contribute to the main issues of activation function outlined in most literature include the vanishing and exploding gradients\unskip~\cite{296343:6576715,296343:6576753} among others, and to remedy these issues, an understanding of the activation functions, which is one of the parameters used in neural network computation, alongside other hyperparameters drives our interest, since these functions are important for better learning and generalisation. The hyperparameters of the neural network are variables of the networks that are set, prior to the optimisation of the network. Other examples of typical hyperparameters of the network include filter size, learning rate, strength of regularisation and position of quantisation level\unskip~\cite{296343:7532566}, with Bergstra et al. outlining that these parameters affect the overall performance of the network.

 However, these advances in configuration of the DL architectures brings new challenges specially to select the right activation functions to perform in different domains from object classification, segmentation, scene description, machine translation, cancer detection, weather forecast, self-driving cars and other adaptive systems to mention a few. With these challenges in mind, the comparison of the current trends in the application of AFs used in DL, portrays a gap in literature as the only similar research paper compared the AFs used in general regression and classification problems in ANNs, with the results reported in Turkish language, which makes it difficult to understand the research results by non Turkish scholars\unskip~\cite{296343:7564721}.

We summarize the trends in the application of existing AFs used in DL and highlight our findings as follows. This compilation is organized into six sections with the first four sections introducing deep learning, activation functions, the summary of AFs used in deep learning and the application trends of AFs in deep architectures respectively. Section five discusses these functions and section six provides the conclusion, alongside some future research direction.
    
\section{Activation Functions}
Activation functions are functions used in neural networks to computes the weighted sum of input and biases, of which is used to decide if a neuron can be fired or not. It manipulates the presented data through some gradient processing usually gradient descent and afterwards produce an output for the neural network, that contains the parameters in the data. These AFs are often referred to as a transfer function in some literature. 

Activation function can be either linear or non-linear depending on the function it represents, and are used to control the outputs of out neural networks, across different domains from object recognition and classification\unskip~\cite{296343:6576743,296343:6577207,296343:6576733,296343:7034152}, to speech recognition\unskip~\cite{296343:6576758,296343:6577215}, segmentation\unskip~\cite{296343:6576712,296343:6577217}, scene understanding and description\unskip~\cite{296343:6576741,296343:6576756}, machine translation\unskip~\cite{296343:6576767,296343:6577221} test to speech systems\unskip~\cite{296343:6576711,296343:6576755}, cancer detection systems\unskip~\cite{296343:6612883,296343:6576768,296343:6576720}, finger print detection\unskip~\cite{296343:7034806,296343:7034805}, weather forecast\unskip~\cite{296343:6576731,296343:6576735}, self-driving cars\unskip~\cite{296343:6577229,296343:6576716}, and other domains to mention a few, with early research results by\unskip~\cite{296343:6576714}, validating categorically that a proper choice of activation function improves results in neural network computing.

For a linear model, a linear mapping of an input function to an output, as performed in the hidden layers before the final prediction of class score for each label is given by the affine transformation in most cases\unskip~\cite{296343:7284218}. The input vectors x  transformation is given by

$\style{font-size:14px}{f(x)=\;\;w^{T}x\;\;+\;b\;\;-(1.1)} $

where x = input, w = weights, b = biases. 

Furthermore, the neural networks produce linear results from the mappings from equation (1.1) and the need for the activation function arises, first to convert these linear outputs into non-linear output for further computation, especially to learn patterns in data. The output of these models are given by
\begin{eqnarray*}\style{font-family:stix}{\style{font-size:12px}{\;\;y=(w_1\;x_1\;+\;\;w_2\;x_2\;+..+\;\;w_n\;x_n+\;b)\;\;\;-(1.2)}}\end{eqnarray*}
These outputs of each layer is fed into the next subsequent layer for multilayered networks like deep neural networks until the final output is obtained, but they are linear by default. The expected output determines the type of activation function to be deployed in a given network.

However, since the output are linear in nature, the nonlinear activation functions are required to convert these linear inputs to non-linear outputs. These AFs are transfer functions that are applied to the outputs of the linear models to produce the transformed non-linear outputs, ready for further processing. The non-linear output after the application of the AF is given by
\begin{eqnarray*}\style{font-family:stix}{\style{font-size:12px}{\;\;y=\alpha(w_1\;x_1\;+\;\;w_2\;x_2\;+..+\;\;w_n\;x_n+\;b)\;\;\;-(1.3)}}\end{eqnarray*}
 Where $\alpha $ is the activation function.

The need for these AFs include to convert the linear input signals and models into non-linear output signals, which aids the learning of high order polynomials beyond one degree for deeper networks. A special property of the non-linear activation functions is that they are differentiable else they cannot work during backpropagation of the deep neural networks\unskip~\cite{296343:7284218}. 

The deep neural network is a neural network with multiple hidden layers and output layer. An understanding of the makeup of the multiple hidden layers and output layer is our interest.  A typical block diagram of a deep learning model is shown in Figure~\ref{figure-fd5c1a80d9ea46ad22d2e539514e674c}, which shows the three layers that make up a DL based system with some emphasis on the positions of activation functions, represented by the dark shaded region in the respective blocks.

\bgroup
\fixFloatSize{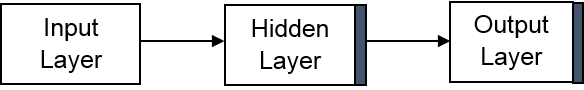}
\begin{figure}[!htbp]
\centering \makeatletter\IfFileExists{images/9d3ffb46-c0c4-4049-b143-d1bd8a592cf3-ublocks.png}{\includegraphics{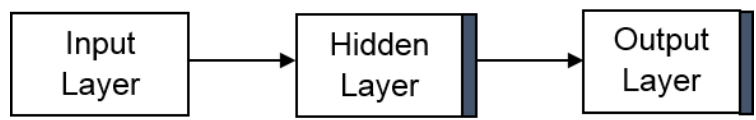}}{}
\makeatother 
\caption{{Block diagram of a DL based system model showing the activation function}}
\label{figure-fd5c1a80d9ea46ad22d2e539514e674c}
\end{figure}
\egroup
The input layer accepts the data for training the neural network which comes in various formats from images, videos, texts, speech, sounds, or numeric data, while the hidden layers are made up of mostly the convolutional and pooling layers, of which the convolutional layers detect the local the patterns and features in data from the previous layers, presented in array-like forms for images while the pooling layers semantically merges similar features into one\unskip~\cite{296343:6576746}. The output layer presents the network results which are often controlled by AFs, specially to perform classifications or predictions, with associated probabilities.

The position of an AF in a network structure depends on its function in the network thus when the AF is placed after the hidden layers, it converts the learned linear mappings into non-linear forms for propagation while in the output layer, it performs predictions. 

The deep architectures are composed of several processing layers with each, involving both linear and non-linear operations, that are learned together to solve a given task. These deeper networks come with better performances though common issues likes vanishing gradients and exploding gradient arises, as a result of the derivative terms which are usually less than 1. With successive multiplication of this derivative terms, the value becomes smaller and smaller and tends to zero, thus the gradient vanishes. Consequently, if the values are greater than 1, successive multiplication will increase the values and the gradient tends to infinity thereby exploding the gradient. Thus, the AFs maintains the values of these gradients to specific limits. These are achieved using different mathematical functions and some of the early proposals of activation functions, used for neural network computing were explored by Elliott, 1993 as he studied the usage of the AFs in neural network\unskip~\cite{296343:6576725}. 

The compilation of the existing activation functions is outlined with the advantages offered by most of the respective functions as highlighted by the authors as found in the literature. 
    
\section{ Summary of Activation Functions}
This section highlights the different types of AFs and their evolution over the years. The AF research and applications in deep architectures, used in different applications has been a core research field to date. The state-of -the -art research results are outlined as follows though, It is worthy to state categorically that this summary of the AFs are not reported in chronological order but arranged with the main functions first, and their improvements following as their variants. These functions are highlighted as follows:

\subsection{Sigmoid Function}The Sigmoid AF is sometimes referred to as the logistic function or squashing function in some literature\unskip~\cite{296343:6576765}. The Sigmoid function research results have produced three variants of the sigmoid AF, which are used in DL applications. The Sigmoid is a non-linear AF used mostly in feedforward neural networks. It is a bounded differentiable real function, defined for real input values, with positive derivatives everywhere and some degree of smoothness\unskip~\cite{296343:6655253}. The Sigmoid function is given by the relationship 
\begin{eqnarray*}\style{font-size:14px}{\;f(x)=\;\left(\;\;\frac1{(1+\;exp^{-x})}\;\right)-(1.4)}\end{eqnarray*}
 The sigmoid function appears in the output layers of the DL architectures, and they are used for predicting probability based output and has been applied successfully in binary classification problems, modeling logistic regression tasks as well as other  neural network domains, with Neal\unskip~\cite{296343:6576752} highlighting the main advantages of the sigmoid functions as, being easy to understand and are used mostly in shallow networks. Moreover, Glorot and Bengio, 2010 suggesting that the Sigmoid AF should be avoided when initializing the neural network from small random weights\unskip~\cite{296343:7519235}.

However, the Sigmoid AF suffers major drawbacks which include sharp damp gradients during backpropagation from deeper hidden layers to the input layers, gradient saturation, slow convergence and non-zero centred output thereby causing the gradient updates to propagate in different directions. Other forms of AF including the hyperbolic tangent function was proposed to remedy some of these drawbacks suffered by the Sigmoid AF.

\subsubsection{Hard Sigmoid Function}The hard sigmoid activation is another variant of the sigmoid activation function and this function is given by 
\begin{eqnarray*}\style{font-size:14px}{\;f(x)=clip\left(\frac{(x+1)}2\;,0,1\right)\;-(1.5)}\end{eqnarray*}
The equation(1.5) can be re-written in the form  
\begin{eqnarray*}\style{font-size:14px}{f(x)\;=\;max\;\left(0,min\left(1,\frac{(x+1)}2\right)\right)-(1.6)}\end{eqnarray*}
A comparison of the hard sigmoid with the soft sigmoid shows that the hard sigmoid offer lesser computation cost when implemented both in a specialized hardware or software form as outlined\unskip~\cite{296343:6577234}, and the authors highlighted that it showed some promising results on DL based binary classification tasks.

\subsubsection{Sigmoid-Weighted Linear Units (SiLU)}The Sigmoid-Weighted Linear Units is a reinforcement learning based approximation function. The SiLU was proposed by Elfwing et al., 2017 and the SiLU function is computed as Sigmoid multiplied by its input\unskip~\cite{296343:6637341}. The AF $a_k $ of a SiLU is given by
\begin{eqnarray*}a_k\left(s\right)\;=\;z_k\alpha\left(z_k\right)\;-(1.7)\end{eqnarray*}
where s = input vector, $z_k $= input to hidden units k. The input to the hidden layers is given by 
\begin{eqnarray*}\style{font-size:14px}{z_k\;=\;\underset i{\sum w_{ik}s_i\;\;+\;b_k}\;\;-(1.8)\;}\end{eqnarray*}
Where $b_k $ is the bias and $w_{ik} $ is the weight connecting to the hidden units k respectively.

The SiLU function can only be used in the hidden layers of the deep neural networks and only for reinforcement learning based systems. The authors also reported that the SiLU outperformed the ReLU function as seen in the response in Figure~\ref{figure-eec3bd8b0d76a13f0505b3be8332e27b}.

\subsubsection{Derivative of Sigmoid-Weighted Linear Units (dSiLU)}The derivative of the Sigmoid-Weighted Linear Units is the gradient of the SiLU function and referred to as dSiLU. The dSiLU is used for gradient-descent learning updates for the neural network weight parameters, and the dSiLU is given by 
\begin{eqnarray*}\style{font-size:14px}{a_k(s)\;=\;\alpha\left(z_k\right)\left(1\;+\;z_k\left(1-\alpha\left(z_k\right)\right)\right)\;-(1.9)}\end{eqnarray*}
The dSiLU function response looks like an overshooting Sigmoid function as shown in Figure~\ref{figure-eec3bd8b0d76a13f0505b3be8332e27b}. The authors highlighted that the dSiLU outperformed the standard Sigmoid function significantly\unskip~\cite{296343:6637341}. 
\bgroup
\fixFloatSize{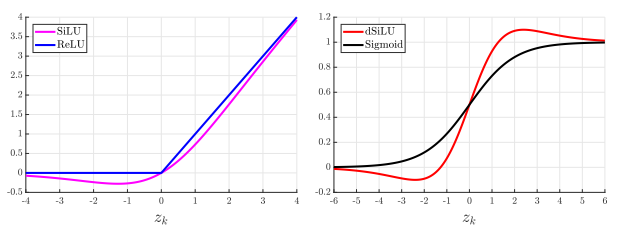}
\begin{figure}[!htbp]
\centering \makeatletter\IfFileExists{images/50b11ab3-1b45-4f8e-86fa-fabe6d2667db-usilu.PNG}{\includegraphics{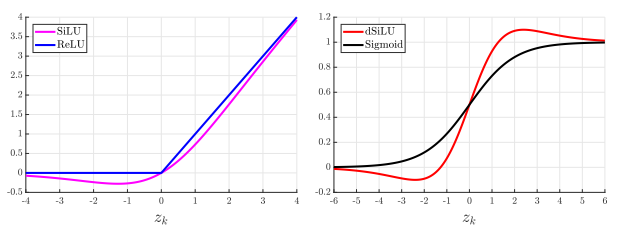}}{}
\makeatother 
\caption{{SiLU response comparison\unskip~\protect\cite{296343:6637341}}}
\label{figure-eec3bd8b0d76a13f0505b3be8332e27b}
\end{figure}
\egroup

\subsection{Hyperbolic Tangent Function (Tanh)}The hyperbolic tangent function is another type of AF used in DL and it has some variants used in DL applications.  The hyperbolic tangent function known as tanh function, is a smoother\unskip~\cite{296343:6576746} zero-centred function whose range lies between -1 to 1, thus the output of the tanh function is given by 
\begin{eqnarray*}\style{font-size:14px}{f(x)\;=\;\;\left(\frac{\;e^{x}-\;e^{-x}\;\;}{e^{x}\;+\;e^{-x}}\right)-(1.10)}\end{eqnarray*}
The tanh function became the preferred function compared to the sigmoid function in that it gives better training performance for multi-layer neural networks\unskip~\cite{296343:6576714,296343:6576752}. However, the tanh function could not solve the vanishing gradient problem suffered by the sigmoid functions as well. The main advantage provided by the function is that it produces zero centred output thereby aiding the back-propagation process. 

A property of the tanh function is that it can only attain a gradient of 1, only when the value of the input is 0, that is when x is zero.  This makes the tanh function produce some dead neurons during computation. The dead neuron is a condition where the activation weight, rarely used as a result of zero gradient. This limitation of the tanh function spurred further research in activation functions to resolve the problem, and it birthed the rectified linear unit (ReLU) activation function.

The tanh functions have been used mostly in recurrent neural networks for natural language processing\unskip~\cite{296343:6943674} and speech recognition tasks\unskip~\cite{296343:6576751}.
\bgroup
\fixFloatSize{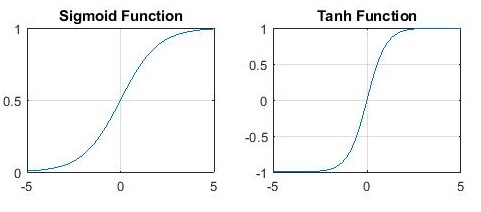}
\begin{figure}[!htbp]
\centering \makeatletter\IfFileExists{images/5ca2fe74-af13-4278-994b-02c6c356ad0f-ufunc.jpg}{\includegraphics{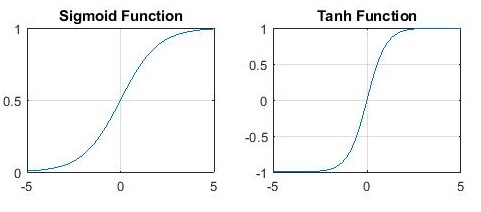}}{}
\makeatother 
\caption{{Pictorial representation of Sigmoid and Tanh activation function responses}}
\label{figure-f6ebf90836878ef932ed48640c2fb7a3}
\end{figure}
\egroup

\subsubsection{Hard Hyperbolic Function}The hard hyperbolic function known as the Hardtanh function is another variant of the tanh activation function used in deep learning applications. The Hardtanh represents a cheaper and more computational efficient version of tanh. The Hardtanh function lies within the range of -1 to 1 and it is given by 
\begin{eqnarray*}\style{font-size:14px}{f(x)=\;\begin{pmatrix}-1\;\;\;\;\;\;\;\;\;\;\;\;\;\;if\;x<-1\\x\;\;if\;-1=\;x\;\leqslant1\\1\;\;\;\;\;\;\;\;\;\;\;\;\;\;if\;x\;>\;1\end{pmatrix}\;-(1.11)}\end{eqnarray*}
 The hardtanh function has been applied successfully in natural language processing\unskip~\cite{296343:6577236}, with the authors reporting that it provided both speed and accuracy improvements.

\subsection{Softmax Function}The Softmax function is another types of activation function used in neural computing. It is used to compute probability distribution from a vector of real numbers. The Softmax function produces an output which is a range of values between 0 and 1, with the sum of the probabilities been equal to 1. The Softmax function\unskip~\cite{296343:7284218} is computed using the relationship
\begin{eqnarray*}\style{font-size:12px}{f(x_i)\;=\;\frac{exp\;\left(x_i\right)}{\displaystyle\;\sum_jexp\left(x_j\right)}\;\;-(1.12)}\end{eqnarray*}
The Softmax function is used in multi-class models where it returns probabilities of each class, with the target class having the highest probability. The Softmax function mostly appears in almost all the output layers of the deep learning architectures, where they are used\unskip~\cite{296343:6576743,296343:6576712,296343:6576749}.

The main difference between the Sigmoid and Softmax AF is that the Sigmoid is used in binary classification while the Softmax is used for multivariate classification tasks.

\subsection{Softsign}The Softsign is another type of AF that is used in neural network computing. The Softsign function was introduced by Turian et al., 2009 and the Softsign is another non-linear AF used in DL applications\unskip~\cite{296343:6576765}. The Softsign function is a quadratic polynomial, given by 
\begin{eqnarray*}\style{font-size:14px}{f(x)=\;\left(\frac x{\left|x\right|\;+\;1}\right)\;\;-(1.13)}\end{eqnarray*}
Where $\left|x\right| $ = absolute value of the input.

The main difference between the Softsign function and the tanh function is that the Softsign converges in polynomial form unlike the tanh function which converges exponentially.

The Softsign has been used mostly in regression computation problems\unskip~\cite{296343:6576745} but has also been applied to DL based test to speech systems\unskip~\cite{296343:6576755}, with the authors reporting some promising results using the Softsign function.

\subsection{Rectified Linear Unit (ReLU) Function}The rectified linear unit (ReLU) activation function was proposed by Nair and Hinton 2010, and ever since, has been the most widely used activation function for deep learning applications with state-of-the-art results to date\unskip~\cite{296343:6576766}. The ReLU is a faster learning AF\unskip~\cite{296343:6576746}, which has proved to be the most successful and widely used function\unskip~\cite{296343:6576757}. It offers the better performance and generalization in deep learning compared to the Sigmoid and tanh activation functions\unskip~\cite{296343:6576773,296343:6576721}. The ReLU represents a nearly linear function and therefore preserves the properties of linear models that made them easy to optimize, with gradient-descent methods\unskip~\cite{296343:7284218}.

The ReLU activation function performs a threshold operation to each input element where values less than zero are set to zero thus the ReLU is given by 
\begin{eqnarray*}\style{font-size:10px}{\;f(x)=\;max\left(0,x\right)=\;\left\{\begin{array}{l}x_{i,\;}\;\;if\;x_i\;\geq\;0\\0,\;\;\;\;if\;x_i<\;0\;\;\end{array}\right.\;\;-(1.14)\;}\end{eqnarray*}
This function rectifies the values of the inputs less than zero thereby forcing them to zero and eliminating the vanishing gradient problem observed in the earlier types of activation function. The ReLU function has been used within the hidden units of the deep neural networks with another AF, used in the output layers of the network with typical examples found in object classification\unskip~\cite{296343:6576743,296343:6576733} and speech recognition applications\unskip~\cite{296343:6576751}.

The main advantage of using the rectified linear units in computation is that, they guarantee faster computation since it does not compute exponentials and divisions, with overall speed of computation enhanced\unskip~\cite{296343:6576773}. Another property of the ReLU is that it introduces sparsity in the hidden units as it squishes the values between zero to maximum. However, the ReLU has a limitation that it easily overfits compared to the sigmoid function although the dropout technique has been adopted to reduce the effect of overfitting of ReLUs and the rectified networks improved performances of the deep neural networks\unskip~\cite{296343:6576727}.

The ReLU and its variants have been used in different architectures of deep learning, which include the restricted Boltzmann machines\unskip~\cite{296343:6576766} and the convolutional neural network architectures\unskip~\cite{296343:6576743,296343:6577207,296343:6576733,296343:6576772,296343:7034152}, although (Nair and Hinton, 2010) outlined that the ReLU has been used in numerous architectures because of its simplicity and reliability\unskip~\cite{296343:6576766}.

The ReLU has a significant limitation that it is sometimes fragile during training thereby causing some of the gradients to die. This leads to some neurons being dead as well, thereby causing the weight updates not to activate in future data points, thereby hindering learning as dead neurons gives zero activation\unskip~\cite{296343:7284218}. To resolve the dead neuron issues, the leaky ReLU was proposed.

\subsubsection{Leaky ReLU (LReLU)}The leaky ReLU, proposed in 2013 as an AF that introduce some small negative slope to the ReLU to sustain and keep the weight updates alive during the entire propagation process\unskip~\cite{296343:6576751}. The alpha parameter was introduced as a solution to the ReLUs dead neuron problems such that the gradients will not be zero at any time during training. The LReLU computes the gradient with a very small constant value for the negative gradient \ensuremath{\alpha } in the range of 0.01 thus the LReLU AF is computed as
\begin{eqnarray*}\style{font-size:14px}{f(x)=\alpha x+x\;=\;\left\{\begin{array}{l}\;x\;\;\;\;\;if\;x\;>\;0\\\alpha x\;\;\;if\;x\;\leq\;0\end{array}\right.\;-(1.15)}\end{eqnarray*}
The LReLU has an identical result when compared to the standard ReLU with an exception that it has non-zero gradients over the entire duration thereby suggesting that there no significant result improvement except in sparsity and dispersion when compared to the standard ReLU and tanh functions\unskip~\cite{296343:6576751}. The LReLU was tested on automatic speech recognition dataset.

\subsubsection{Parametric Rectified Linear Units (PReLU)}The parametric ReLU known as PReLU is another variant of the ReLU AF proposed by He et al., 2015, and the PReLU has the negative part of the function, being adaptively learned while the positive part is linear\unskip~\cite{296343:6576733}. The PReLU is given by 
\begin{eqnarray*}\style{font-size:14px}{\;f(x_i)=\;\begin{pmatrix}x_{i\;\;\;,\;\;\;\;}if\;x_i\;>\;0\\a_ix_i\;,\;\;\;\;if\;x_i\;\leq0\end{pmatrix}\;-(1.16)}\end{eqnarray*}
Where $a_{i\;}\; $ is the negative slope controlling parameter and its learnable during training with back-propagation. If the term $a_i\;=\;0 $  the PReLU becomes ReLU. 

The PReLU can be written in compact form as 
\begin{eqnarray*}\style{font-size:10px}{\;f(x_i\;)=max\left(o,\;x_i\right)\;+\;a_i\;\;min\left(0,\;x_i\right)-(1.17)}\end{eqnarray*}
The authors reported that the performance of PReLU was better than ReLU in large scale image recognition and these results from the PReLU was the first to surpass human-level performance on visual recognition challenge\unskip~\cite{296343:6576733}.

\subsubsection{Randomized Leaky ReLU (RReLU)}The randomized leaky ReLU is a dynamic variant of leaky ReLU where a random number sampled from a uniform distribution $U\left(l,u\right) $ is used to train the network. The randomized ReLU is given by 
\begin{eqnarray*}\style{font-size:14px}{\;f(x_i)=\;\begin{pmatrix}x_{ji\;\;\;,\;\;\;\;}if\;x_{ji}\;\;\geq\;0\\a_{ji}x_{ji}\;,\;\;\;\;if\;x_{ji}\;<\;0\end{pmatrix}\;-(1.18)}\end{eqnarray*}
$Where\;a_i\;\sim\;U\left(l,u\right),\;l\;<\;u\;and\;l,u\;\in\left[0,1\right] $

The test phase uses an averaging technique of all $a_{ji} $ as in the training, without the dropout used in the learned parameter to a deterministic value, obtained as $a_{ji\;}=\;\frac{l\;+\;u}2. $ Thus, the real test output is obtained using the following relationship.
\begin{eqnarray*}\style{font-size:14px}{y_{ji}=\;\;\frac{x_{ji}}{\displaystyle\frac{l\;+\;u}2}\;\;-\;(1.19)\;\;}\end{eqnarray*}
The RReLU has been tested on standard classification datasets and compared against the other variants of the ReLU AF and Xu et al., 2015 validated that LReLU, RReLU and PReLU performs better than the ReLU on classification tasks\unskip~\cite{296343:6577264}.

\subsubsection{S-shaped ReLU (SReLU)}The S-shaped ReLU is another variant of the ReLU AF used to learn both convex and non-convex functions, inspired by the laws of neural sciences and psychophysics. This SReLU AF was proposed by Jin et al., 2015, and it consists of three-piece wise linear functions, formulated by four learnable parameters, which are learned during training of the deep neural network using backpropagation\unskip~\cite{296343:6577246}.

The SReLU is defined by the following mapping relationship
\begin{eqnarray*}\style{font-size:8px}{f(x)=\begin{pmatrix}t_i^{r}\;+\;{a^{r}}_i\left(x_i-{t^{r}}_i\right),\;x_i\;\geq\;{t^{r}}_i\\x_{i\;},\;\;\;\;\;\;\;\;\;\;\;\;\;t_i^{r}\;>x_i\;>\;t_i^{l}\\{t^{l}}_i\;+\;{a^{l}}_i\left(x_i-{t^{l}}_i\right),\;x_i\;\leq\;{t^{l}}_i\end{pmatrix}-(1.20)}\end{eqnarray*}
$where\;{t^{l}}_i,\;{t^{r}}_i\;and\;{a^{l}}_i\; $ are learnable parameters of the network and $i $indicates that the SReLU can vary in different channels. The parameter ${a^{r}}_i $ represents slope of the right line with input above the set threshold ${t^{r}}_i\; $ and ${t^{l}}_i $ are thresholds in positive and negative directions respectively.

The authors highlighted that SReLU was tested on some of the award-winning CNN architectures, the Network in Network architecture alongside GoogLeNet, for on image recognition tasks specifically CIFAR-10, ImageNet, and MNIST standard datasets, and it showed improved results, compared to the other AFs\unskip~\cite{296343:6577246}.

\subsection{Softplus Function}The Softplus AF is a smooth version of the ReLU function which has smoothing and nonzero gradient properties, thereby enhancing the stabilization and performance of deep neural network designed with softplus units.

The Softplus was proposed by Dugas et al., 2001, and the Softplus function is a primitive of the sigmoid function, given by the relationship\unskip~\cite{296343:6576724}
\begin{eqnarray*}f(x)=\log(1\;+\;exp^{x}\;)-(1.21)\end{eqnarray*}
The Softplus function has been applied in statistical applications mostly however, a comparison of the Softplus function with the ReLU and Sigmoid functions by Zheng et al., 2015, showed an improved performance with lesser epochs to convergence during training, using the Softplus function\unskip~\cite{296343:6577248}. 

The Softplus has been used in speech recognition systems\unskip~\cite{296343:6577248}\unskip~\cite{296343:6576760}, of which the ReLU and sigmoid functions have been the dominant AFs, used to achieve auto speech recognition.

\subsection{ Exponential Linear Units (ELUs)}The exponential linear units (ELUs) is another type of AF proposed by Clevert et al., 2015, and they are used to speed up the training of deep neural networks. The main advantage of the ELUs is that they can alleviate the vanishing gradient problem by using identity for positive values and also improves the learning characteristics. They have negative values which allows for pushing of mean unit activation closer to zero thereby reducing computational complexity thereby improving learning speed\unskip~\cite{296343:6576717}. The ELU represents a good alternative to the ReLU as it decreases bias shifts by pushing mean activation towards zero during training.

The exponential linear unit (ELU) is given by 
\begin{eqnarray*}\style{font-size:14px}{\;f(x)=\begin{pmatrix}x,\;\;\;\;\;\;\;\;\;\;\;\;\;\;\;\;if\;x\;>\;0\\\alpha\;exp(x)\;-\;1,\;\;if\;x\leq0\end{pmatrix}\;\;-(1.22)}\end{eqnarray*}
The derivative or gradient of the ELU equation is given as 
\begin{eqnarray*}\;f'=\begin{pmatrix}1,\;\;\;\;\;\;\;\;\;\;\;\;\;\;\;if\;x\;>\;0\\f(x)\;+\;\alpha,\;\;if\;x\;\leq\;0\end{pmatrix}\;\;-(1.23)\end{eqnarray*}
Where $\alpha $= ELU hyperparameter that controls the saturation point for negative net inputs which is usually set to 1.0

The ELUs has a clear saturation plateau in it negative regime thereby learning more robust representations, and they offer faster learning and better generalisation compared to the ReLU and LReLU with specific network structure especially above five layers and guarantees state-of-the-art results compared to ReLU variants. However, a critical limitation of the ELU is that the ELU does not centre the values at zero, and the parametric ELU was proposed to address this issue\unskip~\cite{296343:6577251}.

\subsubsection{Parametric Exponential Linear Unit (PELU)}The parametric ELU is another parameterized version of the exponential linear unit (ELUs), which tries to address the zero centre issue found in the ELUs. The PELU was proposed by Trottier et al., 2017, and it uses the PELU in the context of vanishing gradient to provide some gradient-based optimization framework used to reduce bias shifts while maintaining the zero centre of values\unskip~\cite{296343:6577251}. 

The PELU has two additional parameters compared to the ELU and the modified ELU is given by 
\begin{eqnarray*}\;f(x)=\begin{pmatrix}cx,\;\;\;\;\;\;\;\;\;\;\;\;\;\;\;\;if\;x\;>\;0\\\alpha\;exp^\frac xb\;-\;1,\;\;if\;x\leq0\end{pmatrix}\;\;-(1.24)\end{eqnarray*}
 Where a, b, and c {\textgreater} 0 and c causes a change in the slope in the positive quadrant, b controls the scale of the exponential decay, and $\alpha $ controls the saturation in the negative quadrant. 

Solving the equation of the modified ELU by constraining the projection during training given the parametric ELU which is given by the relationship 
\begin{eqnarray*}\;f(x)=\begin{pmatrix}\frac\alpha bx,\;\;\;\;\;\;\;\;\;\;\;\;\;\;\;\;if\;x\;\geq\;0\\\alpha\;exp^\frac xb\;-\;1,\;\;if\;x\;<\;0\end{pmatrix}\;\;-(1.25)\end{eqnarray*}
$when\;\alpha\;and\;b\;>\;0 $. In compact form, the PELU is given by 
\begin{eqnarray*}f(x_i)=max(0,\;x_i\;)+\;a_i\;\;min(0,\;x_i\;)-(1.26)\end{eqnarray*}
Thus, setting the values of a, b, and c = 1 recovers the original ELU activation function.

The PELU promises to be a good option for applications that requires less bias shifts and vanishing gradients like the CNNs.

\subsubsection{Scaled Exponential Linear Units (SELU)}The scaled exponential linear units(SELU) is another variant of the ELUs, proposed by Klambauer et al., 2017. The SELU was introduced as a self-normalising neural network that has a peculiar property of inducing self-normalising properties. It has a close to zero mean and unit variance that converges towards zero mean and unit variance when propagated through multiple layers during network training, thereby making it suitable for deep learning application, and with strong regularisation, learns robust features efficiently\unskip~\cite{296343:6576742}. 

The SELU is given by 
\begin{eqnarray*}\;f(x)=\tau\begin{pmatrix}x,\;\;\;\;\;\;\;\;\;\;\;\;\;\;\;\;if\;x\;>\;0\\\alpha\;exp(x)\;-\;\alpha,\;\;if\;x\leq0\end{pmatrix}\;\;-(1.27)\end{eqnarray*}
Where $\tau $ is the scale factor.  The approximate values of the parameters of the SELU function are $\alpha\approx\;1.6733\;and\;\lambda\approx\;1.0507. $

The SELUs are not affected by vanishing and exploding gradient problems and the authors have reported that they allow the construction of mappings with properties leading to self normalizing neural networks which cannot be derived using ReLU, scaled ReLU, sigmoid, LReLU and even tanh functions. 

The SELU has been applied successfully to classification tasks\unskip~\cite{296343:6576742} alongside some deep genetic mutation computation\unskip~\cite{296343:6576748}.

\subsection{Maxout Function}The Maxout AF is a function where non-linearity is applied as a dot product between the weights of a neural network and data. The Maxout, proposed by Goodfellow et al., 2013, generalizes the leaky ReLU and ReLU where the neuron inherits the properties of ReLU and leaky ReLU where no dying neurons or saturation exist in the network computation\unskip~\cite{296343:6576729}. The Maxout function is given by 
\begin{eqnarray*}\style{font-size:14px}{f(x)=max(w_1^{T}\;x+b_1\;\;,\;{w^{T}}_2\;x\;+b_2\;)-(1.28)}\end{eqnarray*}
Where w = weights, b = biases, T = transpose.

The Maxout function has been tested successfully in phone recognition applications\unskip~\cite{296343:6577255}. 

The major drawback of the Maxout function is that it is computationally expensive as it doubles the parameters used in all neurons thereby increasing the number of parameters to compute by the network.

\subsection{Swish Function}The Swish AF is one of the first compound AF proposed by the combination of the sigmoid AF and the input function, to achieve a hybrid AF. The Swish activation was proposed by Ramachandran et al., 2017, and it uses the reinforcement learning based automatic search technique to compute the function. The properties of the Swish function include smoothness, non-monotonic, bounded below and unbounded in the upper limits. The smoothness property makes the Swish function produce better optimization and generalization results when used in training deep learning architectures\unskip~\cite{296343:6576757}.

The Swish function is given by 
\begin{eqnarray*}\style{font-size:14px}{f(x)=x\;\cdot\;sigmoid(x)\;=\;\frac x{1\;+\;e^{-x}}-(1.29)}\end{eqnarray*}
 The authors highlighted that the main advantages of the Swish function is the simplicity and improved accuracy as the Swish does not suffer vanishing gradient problems but provides good information propagation during training and reported that the Swish AF outperformed the ReLU activation function on deep learning classification tasks.

\subsection{ELiSH }The Exponential linear Squashing AF known as the ELiSH function is one of the most recent AF, proposed by Basirat and Roth, 2018. The ELiSH shares common properties with the Swish function. The ELiSH function is made up of the ELU and Sigmoid functions and it is given by 
\begin{eqnarray*}\style{font-size:14px}{{f(x)\;=\left\{\begin{array}{l}\left(\frac x{1+\;e^{-x}}\right),\;\;x\;\geq\;0\\\left(\frac{e^{x}\;-\;1}{1+\;e^{-x}}\right),\;\;x\;<\;0\end{array}\right.\;}\;-\;(1.30)}\end{eqnarray*}
The properties of the ELiSH function varies in both the negative and positive parts as defined by the limits. The Sigmoid part of the ELiSH function improves information flow while the Linear parts eliminates the vanishing gradient issues. The ELiSH function has been applied successfully on ImageNet dataset using different deep convolutional architectures\unskip~\cite{296343:6627567}.
\bgroup
\fixFloatSize{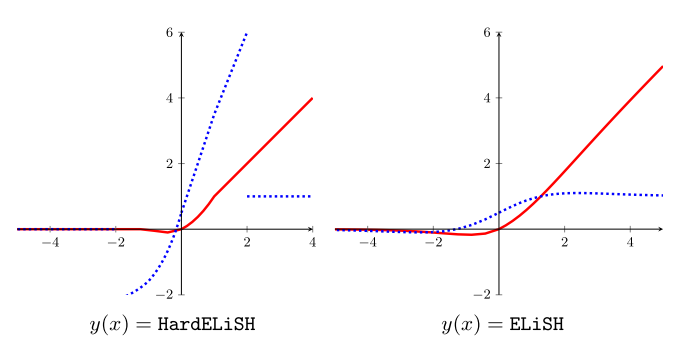}
\begin{figure}[!htbp]
\centering \makeatletter\IfFileExists{images/d0c7001d-42fa-425d-a527-b2b29dfce0c3-uelish.PNG}{\includegraphics{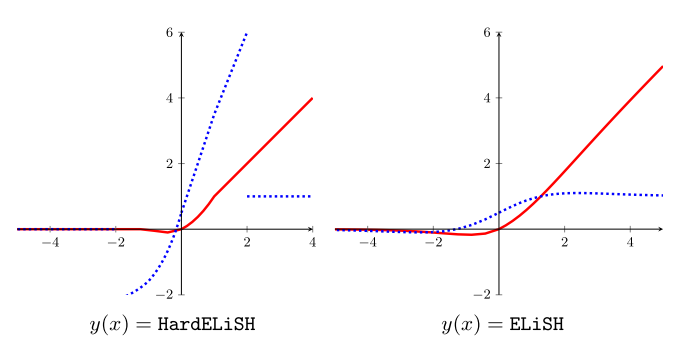}}{}
\makeatother 
\caption{{The ELiSH and HardELiSH function responsees\unskip~\protect\cite{296343:6627567}}}
\label{figure-d31da03aedecde3dc0f2732ba50734b1}
\end{figure}
\egroup

\subsubsection{HardELiSH}The HardELiSH is the hard variant of the ELiSH activation function. The HardELiSH is a multiplication of the HardSigmoid and ELU in the negative part and a multiplication of the Linear and the HardSigmoid in the positive part\unskip~\cite{296343:6627567}. The HardELiSH is given by
\begin{eqnarray*}\style{font-size:12px}{{f(x)=\left\{\begin{array}{l}x\times max\left(0,min\left(1,\left(\frac{x+1}2\right)\right)\right),\;\;x\geq0\\\left(e^{x}-1\right)\times max\left(0,min\left(1,\left(\frac{x+1}2\right)\right)\right),\;x<0\end{array}\right.}-(1.31)}\end{eqnarray*}
The HardELiSH function was tested on ImageNet classification dataset.
    
\section{Comparison of the Trends in Activation Functions Used in Deep Learning Architectures}
The search for the AFs are based on the published research results of the winners of ImageNet Image Large Scale Visual recognition Challenge (ILSVRC) competitions alongside some cited research output from the AF research results found in literature. The ImageNet competition was chosen for trend comparison because it is the competition that produced the first deep learning success\unskip~\cite{296343:6576743}. These forms the core basis for both the search of the functions outlined in this paper and the current trends in the use of AFs, although the scope of the trends goes beyond image recognition as other applications that birthed new DL AFs were included like the natural language processing\unskip~\cite{296343:6576765}

The Image Large Scale Visual Recognition Challenge (ILSVRC) also known as ImageNet is a database of images used for visual recognition competitions. The ImageNet competition is an annual competition where researchers and their teams evaluate developed algorithms on specific datasets, to review improvements in achieved accuracy in visual recognition challenges.

The deep learning architectures are the architectures that has more than one hidden layer, often referred to as multilayer perceptron. These architectures are numerous which include deep feedforward neural networks, convolutional neural networks, long short term memory, recurrent neural networks and the deep generative models like deep Boltzmann machines, deep belief networks, generative adversarial networks and so on\unskip~\cite{296343:6576746,296343:7284218}. The use of the architectures of DL includes to learn patterns in data, map some input function to outputs and many more, and these can only be achieved with specialized architectures.

The AFs used in DL architectures dates back from the beginning of the adoption of the deeper architectures for neural network computations. Prior to 2012, image recognition challenges used shallow architectures and had few AFs embedded in them\unskip~\cite{296343:6577204}. The adoption of the deeper architectures for neural computing was first explored by Krizhevsky et al., 2012 in the Image Large Scale Visual Recognition Challenge (ILSVRC)\unskip~\cite{296343:6576743}, which is an annual event that started in 2010. Since then, the use of the deeper architectures has witnessed huge research advances in optimization techniques for training the deeper architectures, since the deeper the network, the more difficult to train and optimize\unskip~\cite{296343:6577207,296343:6576759}, though the better the performance guarantee.

The AF is a key component for training and optimization of neural networks,  implemented on different layers of DL architectures, is used across domains including natural language processing, object detection, classification and segmentation etc. 

The trends of DL architectures that emerged as the winning architectures of ImageNet competition and other remarkable architectures include: AlexNet; winner of ILSVRC 2012\unskip~\cite{296343:6576743}, ZFNet; the 2013 ILSVRC competition winner\unskip~\cite{296343:6576772}, NiN; the Network in Network architecture\unskip~\cite{296343:6576749}, GoogleNet; the winner of 2015 ILSVRC classification challenge\unskip~\cite{296343:6577207}, VGG16 and VGG19 architectures which emerged as second in classification, winner in localization tasks in 2015 ILSVRC\unskip~\cite{296343:6576761}, Network in Network(NiN)\unskip~\cite{296343:6576749}, ResNet; the 2016 winner of both the ILSVRC \& COCO competitions\unskip~\cite{296343:6576734}, Squeeze and Excitation Network(SeNet); the latest winner of the ILSVRC 2017 competition\unskip~\cite{296343:6576726} alongside SegNet\unskip~\cite{296343:6576712}, DenseNet\unskip~\cite{296343:6576737}, FractalNet\unskip~\cite{296343:6576744}, ResNeXt architecture\unskip~\cite{296343:6576770}, and the modified networks like SqueezeNet architectures\unskip~\cite{296343:6576728}, which had multiple AFs embedded in it.

The Table~\ref{table-wrap-76595f6be84156fda27f602596f78ba8} highlights some of the state-of-the-art architectures of deep neural networks that emerged as a significant improvement to the existing architectures, used for large scale image recognition challenge (ILSVRC), showing the positions of the activation functions used in those architectures.

\begin{table}[!htbp]
\caption{{ Types and positions of AFs used in DL Architectures.} }
\label{table-wrap-76595f6be84156fda27f602596f78ba8}
\def\arraystretch{1}
\ignorespaces 
\centering 
\begin{tabulary}{\linewidth}{p{\dimexpr.20330000000000002\linewidth-2\tabcolsep}p{\dimexpr.19979999999999997\linewidth-2\tabcolsep}p{\dimexpr.13760000000000005\linewidth-2\tabcolsep}p{\dimexpr.4593\linewidth-2\tabcolsep}}
\hline 
Architecture & Hidden Layers & Output\mbox{}\protect\newline Layer & Cross Reference\\
\hline 
AlexNet &
  ReLU &
  Softmax &
  Krizhevsky et al., 2012\\
NiN &
  No activation &
  Softmax &
  Lin et al., 2013\\
ZFNet &
  ReLU &
  Softmax &
  Zeiler \& Fergus, 2013\\
VGGNet &
  ReLU &
  Softmax &
  Simonyan \& Zisserman, 2015\\
SegNet &
  ReLU &
  Softmax &
  Badrinarayanan et al., 2015\\
GoogleNet &
  ReLU &
  Softmax &
  Szegedy et al., 2015\\
SqueezeNet &
  ReLU &
  Softmax &
  Golkov et al., 2016\\
ResNet &
  ReLU &
  Softmax &
  He et al., 2016\\
ResNeXt &
  ReLU &
  Softmax &
  Xie et al., 2017\\
MobileNets &
  ReLU &
  Softmax &
  Howard et al., 2017\\
SeNet &
  ReLU &
  Sigmoid &
  Fu et al., 2017\\
\hline 
\end{tabulary}\par 
\end{table}
The AFs used in more recent DL architectures have the ReLU activation function embedded in them which include the DenseNet\unskip~\cite{296343:6576737}, MobileNets, a mobile version of the convolutional networks\unskip~\cite{296343:6630458}, ResNeXt\unskip~\cite{296343:6576770} as well as the softmax function used in FractalNet\unskip~\cite{296343:6576744}, and many some other architectures.

From Table~\ref{table-wrap-76595f6be84156fda27f602596f78ba8}above, it is evident that the ReLU and the Softmax activation functions are the dormant AFs used in practical DL applications. Furthermore, the Softmax function is used in the output layer of most common practice DL applications however, the most recent DL architecture used the sigmoid function to achieve it prediction at the output layer, while the ReLU units are used in the hidden layers.

In choosing the state-o-the-art architecture for recognition tasks, it is handy from Table~\ref{table-wrap-76595f6be84156fda27f602596f78ba8} to select the current and most recent architecture and understand the architectural make-up  of the network. The SeNet architecture is the current state-of-the-art architecture for recognition tasks as found in literature.

A summary table of all the discussed AFs used in DL is shown in Table~\ref{table-wrap-999bc3969739e652a23ce11e1f02ea98}, with their respective equations, used in computation of these AFs.
\begin{table}[!htbp]
\caption{{DL Activation functions and their corresponding equations for computation.} }
\label{table-wrap-999bc3969739e652a23ce11e1f02ea98}
\def\arraystretch{1}
\ignorespaces 
\centering 
\begin{tabulary}{\linewidth}{p{\dimexpr.06710000000000001\linewidth-2\tabcolsep}p{\dimexpr.21570000000000007\linewidth-2\tabcolsep}p{\dimexpr.7172\linewidth-2\tabcolsep}}
\hline 
S/N & Function & Computation Equation\\
\hline 
1 &
  Sigmoid &
  $\;f(x)=\;\left(\;\;\frac1{(1+\;exp^{-x})}\;\right) $\\
2 &
  HardSigmoid &
  $f(x)\;=\;max\;\left(0,min\left(1,\frac{(x+1)}2\right)\right) $\\
3 &
  SiLU &
  $a_k\left(s\right)\;=\;z_k\alpha\left(z_k\right)\; $\\
4 &
  dSiLU &
  ${\style{font-size:14px}a}_\style{font-size:14px}k\style{font-size:14px}(\style{font-size:14px}s\style{font-size:14px})\style{font-size:14px}\;\style{font-size:14px}=\style{font-size:14px}\;\style{font-size:14px}\alpha\left({\style{font-size:14px}z}_\style{font-size:14px}k\right)\left(\style{font-size:14px}1\style{font-size:14px}\;\style{font-size:14px}+\style{font-size:14px}\;{\style{font-size:14px}z}_\style{font-size:14px}k\left(\style{font-size:14px}1\style{font-size:14px}-\style{font-size:14px}\alpha\left({\style{font-size:14px}z}_\style{font-size:14px}k\right)\right)\right) $\\
5 &
  Tanh &
  $\style{font-size:14px}f\style{font-size:14px}(\style{font-size:14px}x\style{font-size:14px})\style{font-size:14px}\;\style{font-size:14px}=\style{font-size:14px}\;\style{font-size:14px}\;\left(\frac{\style{font-size:14px}\;\style{font-size:14px}e^\style{font-size:14px}x\style{font-size:14px}-\style{font-size:14px}\;\style{font-size:14px}e^{\style{font-size:14px}-\style{font-size:14px}x}\style{font-size:14px}\;\style{font-size:14px}\;}{\style{font-size:14px}e^\style{font-size:14px}x\style{font-size:14px}\;\style{font-size:14px}+\style{font-size:14px}\;\style{font-size:14px}e^{\style{font-size:14px}-\style{font-size:14px}x}}\right) $\\
6 &
  Hardtanh &
  $\style{font-size:14px}f\style{font-size:14px}(\style{font-size:14px}x\style{font-size:14px})\style{font-size:14px}=\style{font-size:14px}\;\begin{pmatrix}\style{font-size:14px}-\style{font-size:14px}1\style{font-size:14px}\;\style{font-size:14px}\;\style{font-size:14px}\;\style{font-size:14px}\;\style{font-size:14px}\;\style{font-size:14px}\;\style{font-size:14px}\;\style{font-size:14px}\;\style{font-size:14px}\;\style{font-size:14px}\;\style{font-size:14px}\;\style{font-size:14px}\;\style{font-size:14px}\;\style{font-size:14px}\;\style{font-size:14px}i\style{font-size:14px}f\style{font-size:14px}\;\style{font-size:14px}x\style{font-size:14px}<\style{font-size:14px}-\style{font-size:14px}1\\\style{font-size:14px}x\style{font-size:14px}\;\style{font-size:14px}\;\style{font-size:14px}i\style{font-size:14px}f\style{font-size:14px}\;\style{font-size:14px}-\style{font-size:14px}1\style{font-size:14px}=\style{font-size:14px}\;\style{font-size:14px}x\style{font-size:14px}\;\style{font-size:14px}\leqslant\style{font-size:14px}1\\\style{font-size:14px}1\style{font-size:14px}\;\style{font-size:14px}\;\style{font-size:14px}\;\style{font-size:14px}\;\style{font-size:14px}\;\style{font-size:14px}\;\style{font-size:14px}\;\style{font-size:14px}\;\style{font-size:14px}\;\style{font-size:14px}\;\style{font-size:14px}\;\style{font-size:14px}\;\style{font-size:14px}\;\style{font-size:14px}\;\style{font-size:14px}i\style{font-size:14px}f\style{font-size:14px}\;\style{font-size:14px}x\style{font-size:14px}\;\style{font-size:14px}>\style{font-size:14px}\;\style{font-size:14px}1\end{pmatrix} $\\
7 &
  Softmax &
  $\style{font-size:12px}f\style{font-size:12px}({\style{font-size:12px}x}_\style{font-size:12px}i\style{font-size:12px})\style{font-size:12px}\;\style{font-size:12px}=\style{font-size:12px}\;\frac{\style{font-size:12px}e\style{font-size:12px}x\style{font-size:12px}p\style{font-size:12px}\;\left({\style{font-size:12px}x}_\style{font-size:12px}i\right)}{\style{font-size:12px}{\displaystyle\;\sum_jexp\left(x_j\right)}}\style{font-size:12px}\;\style{font-size:12px}\; $\\
8 &
  Softplus &
  $f(x)=\log(1\;+\;exp^{x}\;) $\\
9 &
  Softsign &
  $\style{font-size:14px}f\style{font-size:14px}(\style{font-size:14px}x\style{font-size:14px})\style{font-size:14px}=\style{font-size:14px}\;\left(\frac{\style{font-size:14px}x}{\left|\style{font-size:14px}x\right|\style{font-size:14px}\;\style{font-size:14px}+\style{font-size:14px}\;\style{font-size:14px}1}\right) $\\
10 &
  ReLU &
  $\style{font-size:10px}\;\style{font-size:10px}f\style{font-size:10px}(\style{font-size:10px}x\style{font-size:10px})\style{font-size:10px}=\style{font-size:10px}\;\style{font-size:10px}m\style{font-size:10px}a\style{font-size:10px}x\left(\style{font-size:10px}0\style{font-size:10px},\style{font-size:10px}x\right)\style{font-size:10px}=\style{font-size:10px}\;\left\{\begin{array}{l}{\style{font-size:10px}x}_{\style{font-size:10px}i\style{font-size:10px},\style{font-size:10px}\;}\style{font-size:10px}\;\style{font-size:10px}\;\style{font-size:10px}i\style{font-size:10px}f\style{font-size:10px}\;{\style{font-size:10px}x}_\style{font-size:10px}i\style{font-size:10px}\;\style{font-size:10px}\geq\style{font-size:10px}\;\style{font-size:10px}0\\\style{font-size:10px}0\style{font-size:10px},\style{font-size:10px}\;\style{font-size:10px}\;\style{font-size:10px}\;\style{font-size:10px}\;\style{font-size:10px}i\style{font-size:10px}f\style{font-size:10px}\;{\style{font-size:10px}x}_\style{font-size:10px}i\style{font-size:10px}<\style{font-size:10px}\;\style{font-size:10px}0\style{font-size:10px}\;\style{font-size:10px}\;\end{array}\right.\style{font-size:10px}\; $\\
11 &
  LReLU &
  $\style{font-size:14px}f\style{font-size:14px}(\style{font-size:14px}x\style{font-size:14px})\style{font-size:14px}=\style{font-size:14px}\alpha\style{font-size:14px}x\style{font-size:14px}+\style{font-size:14px}x\style{font-size:14px}\;\style{font-size:14px}=\style{font-size:14px}\;\left\{\begin{array}{l}\style{font-size:14px}\;\style{font-size:14px}x\style{font-size:14px}\;\style{font-size:14px}\;\style{font-size:14px}\;\style{font-size:14px}\;\style{font-size:14px}\;\style{font-size:14px}i\style{font-size:14px}f\style{font-size:14px}\;\style{font-size:14px}x\style{font-size:14px}\;\style{font-size:14px}>\style{font-size:14px}\;\style{font-size:14px}0\\\style{font-size:14px}\alpha\style{font-size:14px}x\style{font-size:14px}\;\style{font-size:14px}\;\style{font-size:14px}\;\style{font-size:14px}i\style{font-size:14px}f\style{font-size:14px}\;\style{font-size:14px}x\style{font-size:14px}\;\style{font-size:14px}\leq\style{font-size:14px}\;\style{font-size:14px}0\end{array}\right. $\\
12 &
  PReLU &
  $\style{font-size:14px}\;\style{font-size:14px}f\style{font-size:14px}({\style{font-size:14px}x}_\style{font-size:14px}i\style{font-size:14px})\style{font-size:14px}=\style{font-size:14px}\;\begin{pmatrix}{\style{font-size:14px}x}_{\style{font-size:14px}i\style{font-size:14px}\;\style{font-size:14px}\;\style{font-size:14px}\;\style{font-size:14px},\style{font-size:14px}\;\style{font-size:14px}\;\style{font-size:14px}\;\style{font-size:14px}\;}\style{font-size:14px}i\style{font-size:14px}f\style{font-size:14px}\;{\style{font-size:14px}x}_\style{font-size:14px}i\style{font-size:14px}\;\style{font-size:14px}>\style{font-size:14px}\;\style{font-size:14px}0\\{\style{font-size:14px}a}_\style{font-size:14px}i{\style{font-size:14px}x}_\style{font-size:14px}i\style{font-size:14px}\;\style{font-size:14px},\style{font-size:14px}\;\style{font-size:14px}\;\style{font-size:14px}\;\style{font-size:14px}\;\style{font-size:14px}i\style{font-size:14px}f\style{font-size:14px}\;{\style{font-size:14px}x}_\style{font-size:14px}i\style{font-size:14px}\;\style{font-size:14px}\leq\style{font-size:14px}0\end{pmatrix}\style{font-size:14px}\; $\\
13 &
  RReLU &
  $\style{font-size:14px}\;\style{font-size:14px}f\style{font-size:14px}({\style{font-size:14px}x}_\style{font-size:14px}i\style{font-size:14px})\style{font-size:14px}=\style{font-size:14px}\;\begin{pmatrix}{\style{font-size:14px}x}_{\style{font-size:14px}j\style{font-size:14px}i\style{font-size:14px}\;\style{font-size:14px}\;\style{font-size:14px}\;\style{font-size:14px},\style{font-size:14px}\;\style{font-size:14px}\;\style{font-size:14px}\;\style{font-size:14px}\;}\style{font-size:14px}i\style{font-size:14px}f\style{font-size:14px}\;{\style{font-size:14px}x}_{\style{font-size:14px}j\style{font-size:14px}i}\style{font-size:14px}\;\style{font-size:14px}\;\style{font-size:14px}\geq\style{font-size:14px}\;\style{font-size:14px}0\\{\style{font-size:14px}a}_{\style{font-size:14px}j\style{font-size:14px}i}{\style{font-size:14px}x}_{\style{font-size:14px}j\style{font-size:14px}i}\style{font-size:14px}\;\style{font-size:14px},\style{font-size:14px}\;\style{font-size:14px}\;\style{font-size:14px}\;\style{font-size:14px}\;\style{font-size:14px}i\style{font-size:14px}f\style{font-size:14px}\;{\style{font-size:14px}x}_{\style{font-size:14px}j\style{font-size:14px}i}\style{font-size:14px}\;\style{font-size:14px}<\style{font-size:14px}\;\style{font-size:14px}0\end{pmatrix}\style{font-size:14px}\; $\\
14 &
  SReLU &
  $\style{font-size:8px}f\style{font-size:8px}(\style{font-size:8px}x\style{font-size:8px})\style{font-size:8px}=\begin{pmatrix}\style{font-size:8px}t_\style{font-size:8px}i^\style{font-size:8px}r\style{font-size:8px}\;\style{font-size:8px}+\style{font-size:8px}\;{\style{font-size:8px}a^\style{font-size:8px}r}_\style{font-size:8px}i\left({\style{font-size:8px}x}_\style{font-size:8px}i\style{font-size:8px}-{\style{font-size:8px}t^\style{font-size:8px}r}_\style{font-size:8px}i\right)\style{font-size:8px},\style{font-size:8px}\;{\style{font-size:8px}x}_\style{font-size:8px}i\style{font-size:8px}\;\style{font-size:8px}\geq\style{font-size:8px}\;{\style{font-size:8px}t^\style{font-size:8px}r}_\style{font-size:8px}i\\{\style{font-size:8px}x}_{\style{font-size:8px}i\style{font-size:8px}\;}\style{font-size:8px},\style{font-size:8px}\;\style{font-size:8px}\;\style{font-size:8px}\;\style{font-size:8px}\;\style{font-size:8px}\;\style{font-size:8px}\;\style{font-size:8px}\;\style{font-size:8px}\;\style{font-size:8px}\;\style{font-size:8px}\;\style{font-size:8px}\;\style{font-size:8px}\;\style{font-size:8px}\;\style{font-size:8px}t_\style{font-size:8px}i^\style{font-size:8px}r\style{font-size:8px}\;\style{font-size:8px}>{\style{font-size:8px}x}_\style{font-size:8px}i\style{font-size:8px}\;\style{font-size:8px}>\style{font-size:8px}\;\style{font-size:8px}t_\style{font-size:8px}i^\style{font-size:8px}l\\{\style{font-size:8px}t^\style{font-size:8px}l}_\style{font-size:8px}i\style{font-size:8px}\;\style{font-size:8px}+\style{font-size:8px}\;{\style{font-size:8px}a^\style{font-size:8px}l}_\style{font-size:8px}i\left({\style{font-size:8px}x}_\style{font-size:8px}i\style{font-size:8px}-{\style{font-size:8px}t^\style{font-size:8px}l}_\style{font-size:8px}i\right)\style{font-size:8px},\style{font-size:8px}\;{\style{font-size:8px}x}_\style{font-size:8px}i\style{font-size:8px}\;\style{font-size:8px}\leq\style{font-size:8px}\;{\style{font-size:8px}t^\style{font-size:8px}l}_\style{font-size:8px}i\end{pmatrix} $\\
15 &
  ELU &
  $\style{font-size:14px}\;\style{font-size:14px}f\style{font-size:14px}(\style{font-size:14px}x\style{font-size:14px})\style{font-size:14px}=\begin{pmatrix}\style{font-size:14px}x\style{font-size:14px},\style{font-size:14px}\;\style{font-size:14px}\;\style{font-size:14px}\;\style{font-size:14px}\;\style{font-size:14px}\;\style{font-size:14px}\;\style{font-size:14px}\;\style{font-size:14px}\;\style{font-size:14px}\;\style{font-size:14px}\;\style{font-size:14px}\;\style{font-size:14px}\;\style{font-size:14px}\;\style{font-size:14px}\;\style{font-size:14px}\;\style{font-size:14px}\;\style{font-size:14px}i\style{font-size:14px}f\style{font-size:14px}\;\style{font-size:14px}x\style{font-size:14px}\;\style{font-size:14px}>\style{font-size:14px}\;\style{font-size:14px}0\\\style{font-size:14px}\alpha\style{font-size:14px}\;\style{font-size:14px}e\style{font-size:14px}x\style{font-size:14px}p\style{font-size:14px}(\style{font-size:14px}x\style{font-size:14px})\style{font-size:14px}\;\style{font-size:14px}-\style{font-size:14px}\;\style{font-size:14px}1\style{font-size:14px},\style{font-size:14px}\;\style{font-size:14px}\;\style{font-size:14px}i\style{font-size:14px}f\style{font-size:14px}\;\style{font-size:14px}x\style{font-size:14px}\leq\style{font-size:14px}0\end{pmatrix}\style{font-size:14px}\; $\\
16 &
  PELU &
  $\;f(x)=\begin{pmatrix}cx,\;\;\;\;\;\;\;\;\;\;\;\;\;\;\;\;if\;x\;>\;0\\\alpha\;exp^\frac xb\;-\;1,\;\;if\;x\leq0\end{pmatrix}\;\; $\\
\rAlignHack 17 &
  \rAlignHack SELU &
  \rAlignHack $\;f(x)=\tau\begin{pmatrix}x,\;\;\;\;\;\;\;\;\;\;\;\;\;\;\;\;if\;x\;>\;0\\\alpha\;exp(x)\;-\;\alpha,\;\;if\;x\leq0\end{pmatrix}\;\; $\\
\rAlignHack 18 &
  \rAlignHack Maxout &
  \rAlignHack $\style{font-size:14px}f\style{font-size:14px}(\style{font-size:14px}x\style{font-size:14px})\style{font-size:14px}=\style{font-size:14px}m\style{font-size:14px}a\style{font-size:14px}x\style{font-size:14px}(\style{font-size:14px}w_\style{font-size:14px}1^\style{font-size:14px}T\style{font-size:14px}\;\style{font-size:14px}x\style{font-size:14px}+{\style{font-size:14px}b}_\style{font-size:14px}1\style{font-size:14px}\;\style{font-size:14px}\;\style{font-size:14px},\style{font-size:14px}\;{\style{font-size:14px}w^\style{font-size:14px}T}_\style{font-size:14px}2\style{font-size:14px}\;\style{font-size:14px}x\style{font-size:14px}\;\style{font-size:14px}+{\style{font-size:14px}b}_\style{font-size:14px}2\style{font-size:14px}\;\style{font-size:14px}) $\\
\rAlignHack 19 &
  \rAlignHack Swish &
  \rAlignHack $\style{font-size:14px}f\style{font-size:14px}(\style{font-size:14px}x\style{font-size:14px})\style{font-size:14px}=\style{font-size:14px}x\style{font-size:14px}\;\style{font-size:14px}\cdot\style{font-size:14px}\;\style{font-size:14px}s\style{font-size:14px}i\style{font-size:14px}g\style{font-size:14px}m\style{font-size:14px}o\style{font-size:14px}i\style{font-size:14px}d\style{font-size:14px}(\style{font-size:14px}x\style{font-size:14px})\style{font-size:14px}\;\style{font-size:14px}=\style{font-size:14px}\;\frac{\style{font-size:14px}x}{\style{font-size:14px}1\style{font-size:14px}\;\style{font-size:14px}+\style{font-size:14px}\;\style{font-size:14px}e^{\style{font-size:14px}-\style{font-size:14px}x}} $\\
\rAlignHack 20  &
  \rAlignHack ELiSH &
  \rAlignHack $\style{font-family:stix}{\style{font-size:11px}{f(x)\;=\left\{\begin{array}{l}\left(\frac x{1+\;e^{-x}}\right),\;\;x\;\geq\;0\\\left(\frac{e^{x}\;-\;1}{1+\;e^{-x}}\right),\;\;x\;<\;0\end{array}\right.\;}} $\\
\rAlignHack 21 &
  \rAlignHack HardELiSH &
  \rAlignHack $\style{font-family:stix}{\style{font-size:12px}f\style{font-size:12px}(\style{font-size:12px}x\style{font-size:12px})\style{font-size:12px}=\left\{\begin{array}{l}\style{font-size:12px}x\style{font-size:12px}\times\style{font-size:12px}m\style{font-size:12px}a\style{font-size:12px}x\left(\style{font-size:12px}0\style{font-size:12px},\style{font-size:12px}m\style{font-size:12px}i\style{font-size:12px}n\left(\style{font-size:12px}1\style{font-size:12px},\left(\frac{\style{font-size:12px}x\style{font-size:12px}+\style{font-size:12px}1}{\style{font-size:12px}2}\right)\right)\right)\style{font-size:12px},\style{font-size:12px}\;\style{font-size:12px}\;\style{font-size:12px}x\style{font-size:12px}\geq\style{font-size:12px}0\\\left(\style{font-size:12px}e^\style{font-size:12px}x\style{font-size:12px}-\style{font-size:12px}1\right)\style{font-size:12px}\times\style{font-size:12px}m\style{font-size:12px}a\style{font-size:12px}x\left(\style{font-size:12px}0\style{font-size:12px},\style{font-size:12px}m\style{font-size:12px}i\style{font-size:12px}n\left(\style{font-size:12px}1\style{font-size:12px},\left(\frac{\style{font-size:12px}x\style{font-size:12px}+\style{font-size:12px}1}{\style{font-size:12px}2}\right)\right)\right)\style{font-size:12px},\style{font-size:12px}\;\style{font-size:12px}x\style{font-size:12px}<\style{font-size:12px}0\end{array}\right.} $\\
\hline 
\end{tabulary}\par 
\end{table}

\section{Discussions}
We have identified the key AFs used in deep learning applications and from the literature, we have outlined that there are four variants of the rectified linear units (ReLU), apart from the original ReLU. Furthermore, the Sigmoid has three variants, two variants for the exponential linear units(ELU) functions, with the Hyperbolic tangent and ELiSH functions having their respective hard versions as their only variants. The Softmax, Softsign, Softplus, Maxout and Swish functions has no variants.

The summary of the various AFs include the Sigmoid with HardSigmoid, SiLU and dSiLU variants, the ELU having the PELU and SELU as its variants, the ReLU with LReLU, PReLU, RReLU and SReLU as its variants. 

Perhaps, the DL architectures have multiple AFs embedded to every given network and it is noteworthy to highlight that these multiple AFs are used at different layers, to perform gradient computations in a single network, thereby obtaining an effective learning and characterisation of the presented inputs.

The linear units of AF are the most studied types of AF with the rectified and exponential variants, providing eight different variants of activation functions, used in DL applications and these eight linear variants include ReLU, LReLU, PReLU, RReLU, SReLU, ELU, PELU and SELU.

A summary of the AFs used across various domains is shown in Table 2 alongside the formulae for the computation of these AFs, used in DL applications.

The ELU's has been highlighted as a faster learning AF compared to their ReLU counterpart\unskip~\cite{296343:6577262,296343:6577251}, and this assertion has been validated by Pedamonti, 2018, after an extensive comparison of some variants of the ELU and ReLU AF on the MNIST recognition dataset\unskip~\cite{296343:6577263}.

The PELU and PReLU are two different parametric AFs which are developed using exponential and linear units but both have learnable parameters. It is also evident that there are other AFs that performs better than the ReLU, which has been the most consistent AF for DL applications since invention\unskip~\cite{296343:6576757}\unskip~\cite{296343:6576733}. Xu et al., 2015 validated that the other variants of the ReLU, including LRELU, PReLU and RReLU performs better than the ReLU but some of these functions lack theoretical justifications to support their state-of-the-art results\unskip~\cite{296343:6577264}.

Furthermore, the parametric AFs have been a development in emerging applications where the AF was used as a learnable parameter from the dataset, thus looks to be a promising approach in the new functions developed most recently as observed in SReLU, PELU and PReLU.

Apart from the ReLU and SiLU AFs that were developed on restricted Boltzmann machines\unskip~\cite{296343:6637341,296343:6576766}, and LReLU developed with neural network acoustic models\unskip~\cite{296343:6576751}, almost all the other AFs were developed on convolutional neural networks and tested on classification datasets. This further demonstrated that deep learning tasks are thriving mostly in classification tasks, using convolutional neural networks.

The most notable observation on the use of AFs for DL applications is that the newer activation functions seem to outperform the older AFs like the ReLU, yet even the latest DL architectures rely on the ReLU function. This is evident in SeNet where the hidden layers had the ReLU activation function and the Sigmoid output\unskip~\cite{296343:6576726}. However, current practices does not use the newly developed state-of-the-art AFs but depends on the tested and proven AFs, thereby underlining the fact that the newer activation functions are rarely used in practice.
    
\section{Conclusion}
This paper provides a comprehensive summary of AFs used in deep learning and most importantly, highlights the current trends in the use of these functions in practice against the state -of-the-art research results, which until now, has not been published in any literature.

We first presented a brief introduction to deep learning and activation functions, and then outlined the different types of activation functions discussed, with some specific applications where these functions were used in the development of deep learning based architectures and systems. 

The AFs have the capability to improve the learning of the patterns in data thereby automating the process of features detection and justifying their use in the hidden layers of the neural networks, and usefulness for classification purposes across domains.

These activation functions have developed over the years with the compounded activation functions looking towards the future of activation functions research. Furthermore, it is also worthy to state that there are other activation functions that have not been discussed in this literature as we focused was on the activation functions, used in deep learning applications. A future work would be to compare all these state-of-th-art functions on the award-winning architectures, using standard datasets to observe if there would be improved performance results.



%

\bibliographystyle{IEEEtran}

\bibliography{\jobname}

\end{document}